# Geometric Convolutional Neural Network

# for Analyzing Surface-Based Neuroimaging Data

Si-Baek Seong[1*], Chongwon Pae[1], and Hae-Jeong Park[1,2,3*]

[1]Brain Korea 21 PLUS Project for Medical Science, Yonsei University College of Medicine, [2]Department of Nuclear Medicine, Radiology, and Psychiatry, Severance Hospital, Yonsei University College of Medicine, [4]Department of Cognitive Science, Yonsei University, Seoul, Republic of Korea.

* equally contributed authors

**Author:**

Si-Baek Seong

Email: sbseong@yuhs.ac, firstsb@gmail.com

**Corresponding Author:**

Hae-Jeong Park, PhD

Professor, Department of Nuclear Medicine, Radiology, Psychiatry,

Yonsei University College of Medicine,

50 Yonsei-ro, Sinchon-dong, Seodaemun-gu, Seoul, Republic of Korea 120-752,

Tel.: 82-2-2228-2363; Fax: 82-2-393-3035

Email: parkhj@yonsei.ac.kr

# Abstract

In the machine learning, one of the most popular deep learning methods is convolutional neural network (CNN), which utilizes shared local filters and hierarchical information processing analogous to the brain's visual system. Despite its popularity in recognizing two-dimensional images, the conventional CNN is not directly applicable to semi-regular geometric mesh surfaces, over which the cerebral cortex is often represented. In order to apply CNN to surface-based brain research, we propose a geometric CNN (gCNN) that deals with data representation over a mesh surface and renders pattern recognition in a multi-shell mesh structure. To make it compatible with the conventional CNN toolbox, gCNN includes data sampling over the surface and a data reshaping method for the convolution and pooling layers. We evaluated the performance of gCNN in sex classification using the cortical thickness maps of both hemispheres in the Human Connectome Project. The classification accuracy of gCNN was significantly higher than those of a support vector machine and a two-dimensional CNN for thickness maps generated by a map projection. gCNN also demonstrated position invariance of local features that renders reuse of its pre-trained model for applications other than the purpose for which the model was trained, without significant distortions in the final outcome. The superior performance of gCNN is attributable to the CNN properties stemming from its brain-like architecture and surface-based representation of the cortical information. gCNN provides much-needed access to surface-based machine learning that can be used in both scientific investigations and clinical applications.

# Introduction

In the machine learning domain, convolutional neural network (CNN) (Krizhevsky et al., 2012; LeCun et al., 1998) has made an enormous impact on pattern recognition. Analogous to the tiled receptive fields in the hierarchical visual system in the brain, this technique utilizes replicated (shared) local filters in a convolution layer, which makes it efficient in detecting common local features, regardless of their position in image space. CNN also takes advantage of the hierarchical architecture by utilizing a pooling layer that represents distributed local features as global patterns. Due to its strength in hierarchical feature detection, CNN has not only been used in image-based pattern recognition but also in identifying patterns in three dimensions; in volume (Kamnitsas et al., 2015; Maturana and Scherer, 2015; Nie et al., 2016), in time (e.g., dynamic images) (Huang et al., 2015; Ji et al., 2013), and in different modalities (Kamnitsas et al., 2015; Nie et al., 2016). Despite many variations, CNN is most popularly used for recognition of patterns in two-dimensional image space. However, the conventional CNN technique cannot be directly applied to data in three-dimensional (3D) geometric surface space, such as cortical thickness on the cortical surface.

In brain research, brain morphometry and functionality are often represented in the cortical surface geometry (Dale et al., 1999; Fischl et al., 1999a; MacDonald et al., 2000; Van Essen and Drury, 1997; Van Essen et al., 1998). The most promising aspect of the surface-based approach is the ability to explore cortical thickness, which can typically be represented on the surface (Fischl and Dale, 2000; Kabani et al., 2001; Kuperberg et al., 2003; Narr et al., 2005). For example, Park et al. (2009) showed the specificity of the surface-based cortical thickness representation compared to volumetric representation. Metabolic activity can also be efficiently processed in the cortical surface (Greve et al., 2014; Park et al., 2006). Despite the many advantages of cortical surface representation of brain structure and function, no efficient method for CNN over the cortical surface has been proposed.

In this paper, we propose a geometric CNN (gCNN) that expresses data representation on a geometric surface and recognizes cortical distribution patterns. Although the method can be expanded over any surface shape, we focus on the spherical surface because of its simplicity. The cortical surface has the same topology as a spherical surface; consequently, the cortical surface is often treated as a spherical surface—for example, in surface-based registration across brains (Fischl et al., 1999b). gCNN performs convolution and pooling over the spherical surface to capture hierarchical features on the surface. In order to demonstrate the performance of the proposed method, we applied gCNN to sex classification using 733 cortical thickness maps of the Human Connectome Project (HCP) (Van Essen et al., 2012). We compared the classification accuracy of gCNN with those of a conventional support vector machine (SVM) and a conventional two-dimensional CNN for thickness maps after projecting cortical thickness into two-dimensional image space (pCNN).

Finally, we evaluated the performance of gCNN in detecting position invariant local features compared to pCNN by testing the reusability of the low-level features after global rotation of thickness distribution at angles of 45° and 90°.

## Methods

*Geometric CNN (gCNN)*

The main units comprising gCNN are surface-based convolution layers and pooling layers. The functions of these layers are similar to those of conventional CNNs, with the exception that they deal with surface data. In order to utilize available conventional CNN toolboxes, we added data reshaping steps to each layer. Fig. 1 illustrates the conceptual architecture and the implemented architecture of gCNN. The architecture comprises an input data layer, mesh convolution layers with data reshaping, batch normalization layers, rectified linear unit (ReLU) layers, mesh pooling layers, and a fully connected layer with a softmax output function.

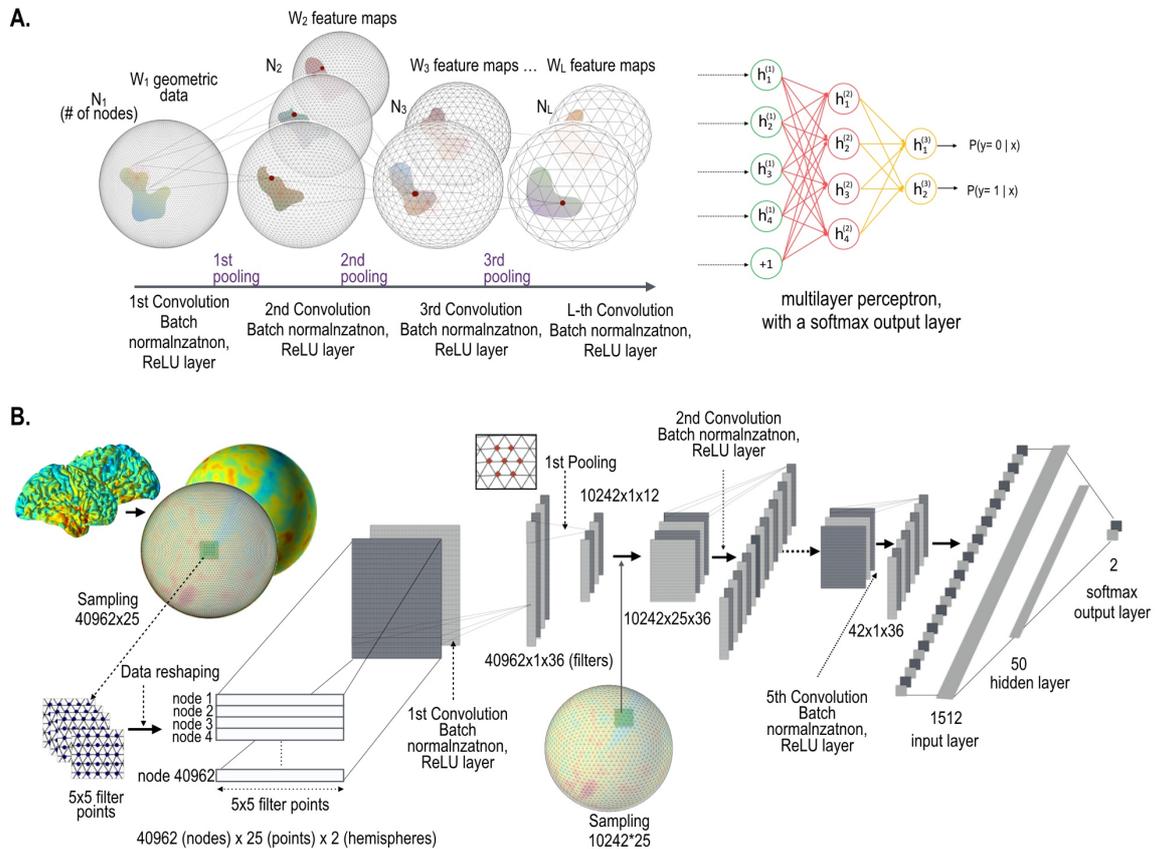

**Fig. 1.** Architecture of gCNN. (A) Conceptual architecture of gCNN. When data on the cortical surface enter the convolution layer with batch normalization and rectified linear unit (ReLU) output function layers, W feature maps (corresponding to the number of filters at each convolution layer) are generated. The dimension of the nodes decreases from N1 to NL as the data pass through the pooling layers. gCNN ends with a multilayer classifier. (B) Implementation level architecture of gCNN. The input data 40962 (nodes) × 25 (filter sample points) × 2 (hemispheres) are convolved with 36 filters, which are reduced to 42 (nodes) × 36 (filtered outputs, i.e., features) after five convolution and pooling steps. As the data pass through the layers, the number of features increases but the dimension of the nodes decreases.

Finally, the convolution-pooling data enter the fully connected multilayer perceptron comprising a hidden layer with 50 nodes and a softmax output layer with two nodes.

*1) The convolution layer*

The convolution filter can be defined according to the patch geometry (i.e., the sampling methods over the mesh) of the filter. For each node in the mesh, the proposed method provides three types of filter according to spatial patch geometry: rectangular, circular, and polygonal patch grids (Fig. 2). Each patch has spatially distributed sampling points (filter points) arranged at regular intervals around the node and subjected to the convolution operation. A rectangular filter processes every patch composed of rectangular points. A circular patch has points centered on a mesh node arranged radially. A polygonal filter convolutes inputs from the neighboring nodes for the target node. To reduce computational loads, the intensity value corresponding to the filter points on the mesh surface is obtained by interpolating the intensity value of the nearest neighbor nodes to which the filter points belong.

In the conventional CNN, each image patch is convoluted with a filter in a sliding window manner. In order to utilize conventional CNN toolboxes using GPUs, we rearranged the sampled filter points to render the convolution operation as a simple filter weighting process (Fig. 2). For each node n, on the surface with a total node number N, filter points sampled from the rectangular, circular, or polygonal grid of the node are first reshaped into a row vector in the full-node filter point matrix **I** (dimension: number of nodes × number of filter points). The output vector $\mathbf{O_f}$ (dimension: number of nodes × 1) is obtained by multiplying the filter point matrix **I** by the f-th mesh filter vector $\mathbf{W}_f$ (dimension: number of filter points × 1), $\mathbf{O_f} = \mathbf{I} \times \mathbf{W}_f$, f = 1, …, Fb (total number of filters at each convolution layer). Fig. 2A shows an example of a rectangular patch for each node, which samples the surface data at rectangular filter points at regular intervals to create the **I**(i, j) matrix (Fig. 2D). We multiply the sampled intensity $\mathbf{I}(i, j)_n$ (or thickness in the current study) obtained for each node by a filter weight vector $\mathbf{W}_f$, which generates an output [$\mathbf{O}_f$(n)] corresponding to node n (Fig. 2G). The filter weight vector $\mathbf{W}_f$ (f = 1, …, Fb) is updated to optimize performance while training gCNN.

A circular patch composed of multiple circles for each node can also be constructed, as shown in Fig. 2B and Fig. 2E. Patches at all nodes construct a full-node filter point matrix, as shown in Fig. 2H. Figs. 2C and 2F show a polygonal patch composed of the first-order and the second-order neighbor nodes (as filter points) of the corresponding node. These different types of patch geometries can be used on a case by case basis. Circular and polygonal patch may be appropriate in some specific applications where the divergence of a node over the surface is important. In the current study, we used the rectangular filter point grid for pattern classification of cortical thickness since our preliminary test showed the best performance with the rectangular filter in the sex classification. However, we can also choose circular and polygonal patches depending on their application.

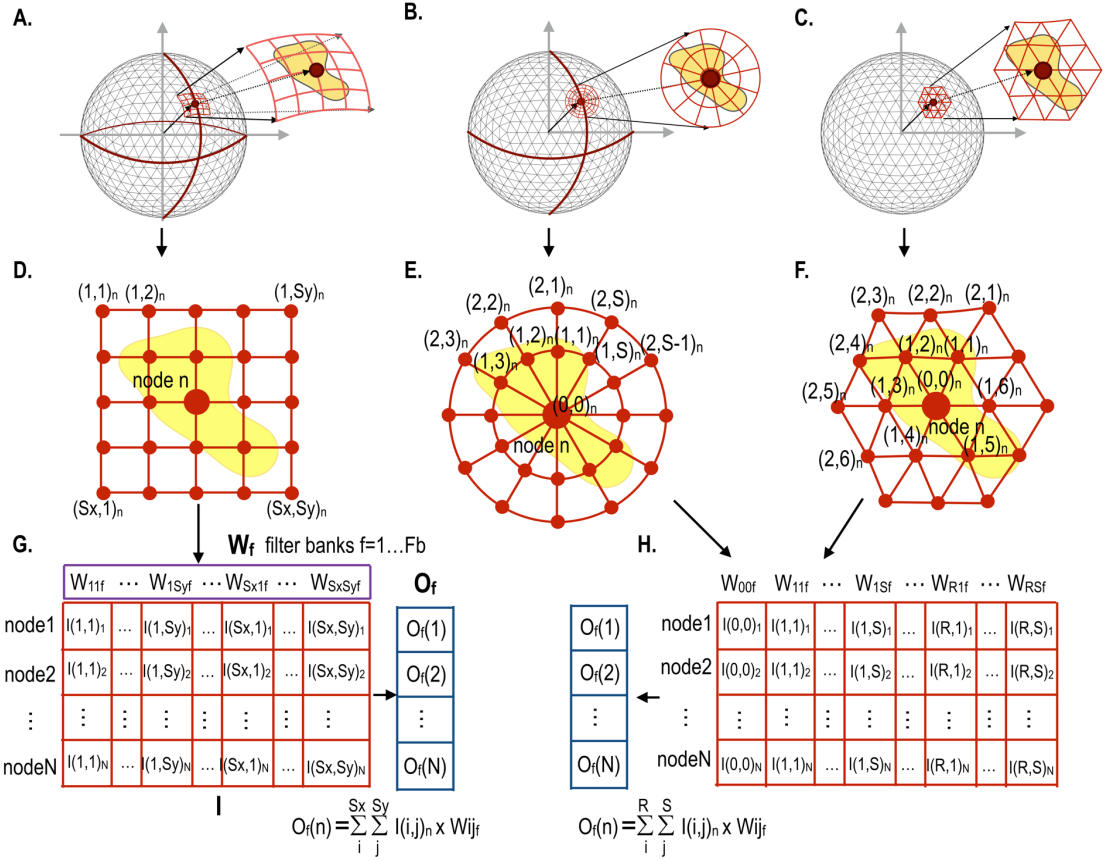

**Fig. 2.** Data sampling and reshaping methods for three different types of filters according to patch geometry for mesh convolution. For all nodes in the sphere (A), rectangular filter points in a patch for each node n ($Sx \times Sy$ points for each node) (D) compose a row vector of a full-node filter point matrix **I** with dimensions [$N \times (Sx \times Sy)$] in (G). A circular filter point matrix (B) can be similarly constructed by sampling circular points over the surface (E). Polygonal filter points in (C) are composed of up to R-th order neighbor nodes (F). A circular patch in (E) and a polygonal patch in (F) at each node compose a full-node filter point matrix I respectively, as shown in (H). The convolution operation can be performed by multiplication of the filter point matrix **I** by a filter weight vector $\mathbf{W}_f$, resulting in output vector $\mathbf{O}_f$ for the filter weight vector f. Fb is the number of filters for each convolution layer. These filtered data are down-sampled in the next pooling layer.

*2) The mesh pooling layer*

The convolution output is subsampled in the subsequent pooling layer (Fig. 3). For pooling, we utilize the regularity characteristic of the icosahedron that can be expanded easily by a simple rule. The icosahedron increases the number of nodes by adding a new node to each center of three triangular edges, which divides one parent triangle into four child triangles. Iterating this process creates a fine spherical surface with node numbers 42, 162, 642, 2562, 10242, 40962, and so on. Subsampling can be performed in the reverse order of icosahedron expansion, leading to the spherical surface of the pre-expansion stage. Fig. 3C shows an example of pooling over the icosahedron spherical system. In the current study, we used a mean pooling over the mesh structure during the forward propagation, which assigns the parent node with an average of the convolution outputs at the child nodes (Fig. 4A). During the error backpropagation, errors in the parent nodes are evenly distributed to their child nodes, as illustrated in Fig. 4B.

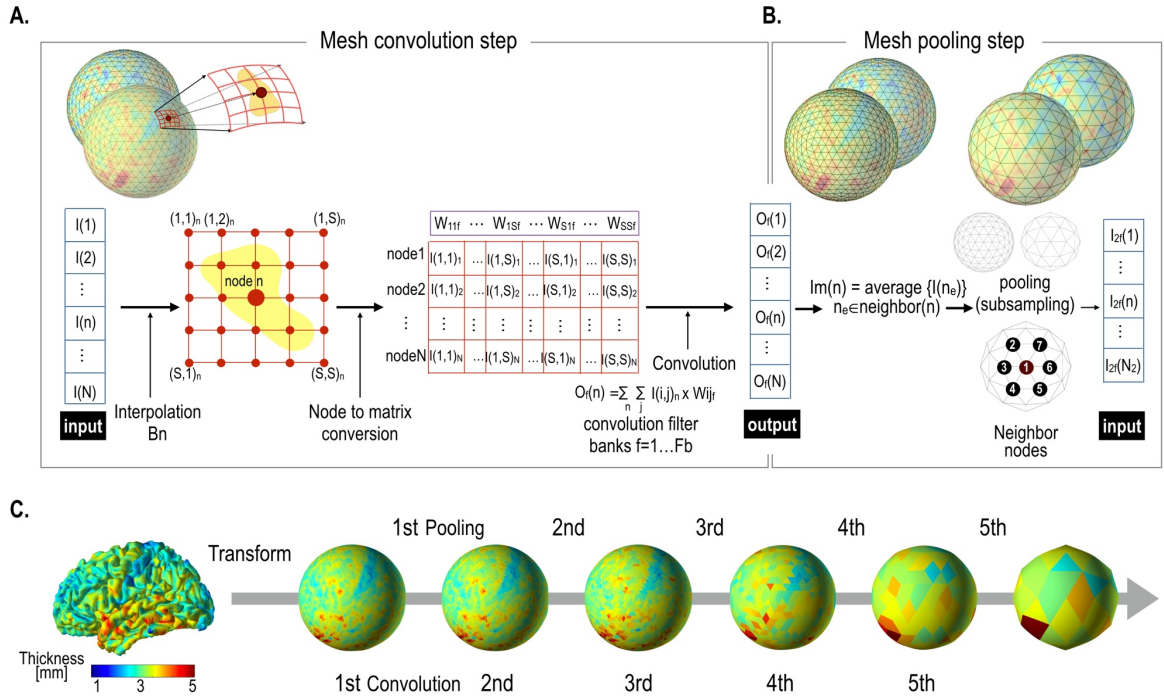

**Fig. 3.** Mesh pooling after mesh convolution operation. (A) A mesh convolution layer processes inputs and generates outputs by multiplication of filter vectors with reshaped thickness data. (B) The output value of the convolution layer is subsampled at the pooling layer. In this study, mean pooling was used. (C) Cortical thickness representation example of successive pooling processes. Cortical thickness data over a realistic cortical sheet (node size = 32492) are transformed to a spherical surface. In order to utilize the regularity characteristic of the icosahedron, we interpolated 32492 nodes into 40962 nodes, which were subsequently subsampled to 10242 ($2^{nd}$), 2562 ($3^{rd}$), 642($4^{th}$), 162($5^{th}$), and 42($6^{th}$) spherical nodes, from the local to the global direction.

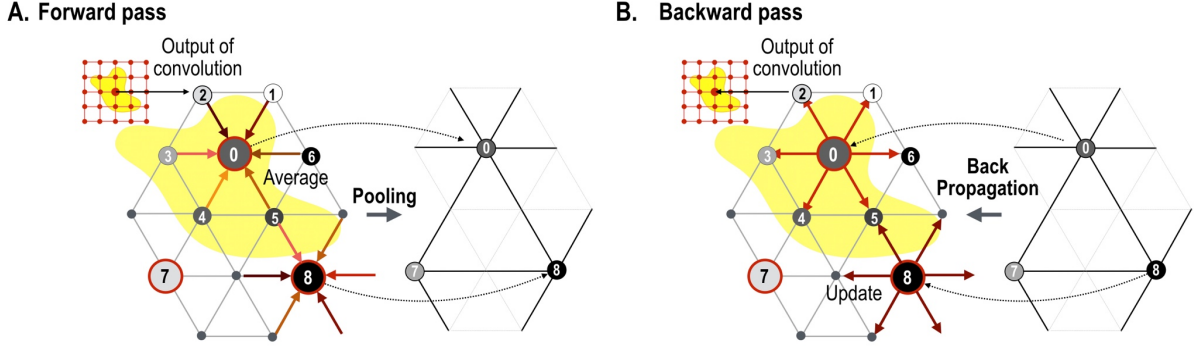

**Fig. 4.** Forward mesh pooling and error backpropagation step. (A) Currently, gCNN uses a mean pooling method that assigns the average value of the neighbor nodes to the target parent node. After pooling, only parent nodes remain (0, 7, 8 in this illustration) and compose a less dense spherical surface. When backpropagation is performed, the error is distributed equally to all the neighboring (child) nodes.

*Implementation of gCNN*

To implement the proposed method, we modified the MatConvNet toolbox (Vedaldi and Lenc, 2015), available from http://www.vlfeat.org/matconvnet/. We constructed gCNN by repeating the mesh convolution and pooling layers. Between the convolution and pooling layers, we inserted a batch normalization unit and a ReLU to increase the training performance. Semi-batch training with a batch size of 50 increases learning performance and expedites network learning, according to Ioffe and Szegedy (2015). The ReLU layer naturally leads to sparse nodal activity (Glorot et al., 2011). We used average pooling, which exhibited better performance than max pooing in preliminary evaluations conducted in the current study. Thus, the main operational complex comprises a convolution layer, a batch normalization layer, a ReLU layer, and a pooling layer. The final unit is a fully connected multilayer perceptron with a softmax output layer. Conventional backpropagation and gradient descent algorithm are used to update the model weights, which are implemented in MatConvNet. To minimize the overfitting problem, we adjusted the learning rate from 0.02 to 0.001 along the training process.

*Application to sex classification*

In order to evaluate the performance of gCNN, we applied gCNN to sex classification using 733 cortical thickness maps in the HCP database. In the HCP database, the cortical thickness is measured over 32492 nodes in each hemispheric cortical surface. We interpolated the 32492 nodal thicknesses into 40962 nodal thicknesses using bilinear interpolation to make the pooling steps simple based on the icosahedron architecture. For each node, we normalized each individual's thickness data by demeaning (i.e., subtracting the average thickness (across entire brain) of the individual from the individual's thickness values). The test data were also normalized by demeaning. In the first convolution layer, we resampled 25 rectangular grid (i.e. a patch) points (5×5) from each node of the surface, which resulted in a full-node filter point matrix comprising 40962 nodes × 25 points thickness data for each hemisphere (Fig. 1). The patch size 5 × 5 was chosen empirically by considering the node resolution and spatial extent of the patch.

Both the left and right hemispheric cortical thickness maps were combined to make an additional dimension. We used 36 convolution filter banks with size [25 × 1] in the first layer. These filter banks were then convolved with a filter point matrix, with a size of 40962 (nodes) × 25 (reshaped thickness filter points) × 2 (hemispheres), as shown in Fig. 1.

*Support vector machine (SVM) of the cortical thickness.*

In order to compare the proposed method with conventional classifiers, we conducted SVM classification using LIBSVM (Chang and Lin, 2011). In order to optimize the SVM, we evaluated five types of SVM classifiers (C-SVC, nu-SVC, one-class SVM, epsilon-SVR, and nu-SVR) with four different kernel types (linear, polynomial, radial basis function, and sigmoid). The best performing classifier and kernel were C-SVC and linear kernel with a kernel regulation parameter C = 1, epsilon: 0.001.

*Conventional CNN for projected cortical thickness images*

Surface-based representation has often been projected into a two-dimensional image, for example, earth maps. Similarly, pattern classification of the cortical thickness can be conducted on the two-dimensional image space after projection. In order to compare gCNN with a conventional CNN of projected thickness images (hereafter, called pCNN), we projected cortical thickness in the spherical surface into the two-dimensional image. Among many two-dimensional projection methods for spherical data, we conducted projection by latitude and longitude. The portion of the non-cortical brain was set to zero in the training and testing processes. All cortical thickness data were projected into 224 × 224 images. In order to utilize continuous information (continuous over the boundary) in the spherical data, we padded marginal five pixels from the other side of the image (the filter size of the first convolution layer was 11 × 11; thus, at least five pixels were needed for convolution). We adopted a CNN architecture from ImageNet-VGG-F (Chatfield et al., 2014), which has six convolution layers (with a normalization layer and an activation layer (ReLU) for each convolution layer) and a fully connected softmax classifier layer. Please see Table 2 in the Appendix for the detailed model.

*Performance evaluation*

In order to evaluate the performance of the classifiers (gCNN, SVM, and pCNN), we divided 733 thickness samples into a training-validation set (670 samples) and a test set (63 samples). Using the training-validation set, we conducted 10-fold cross-validation, by splitting the thickness dataset randomly into 90% for training a model and 10% for validating the model. During the cross-validation, we optimized a model of each fold and evaluated the performance of the trained model using the test set. Based on the ratio of males and females (328 males and 405 females used in this study), the average numbers of males and females in the validation set of a fold (67 samples in each fold) were chosen as 29.6 and 37.3, respectively. The numbers of males and females in the test set were 28 and 35, in a similar male/female ratio as the entire data set.

We trained a model for each fold iteratively for a total epoch (or iteration) size of 40. When model training was not saturated (i.e., the difference in error rates between consecutive epochs is not 0) after 40 epochs, we extended training up to 70 epochs. All gCNN models (a model per a fold) were saturated before 40 epochs in the experiment. Meanwhile, most pCNN models were saturated after 40–70 epochs.

We optimized gCNN and pCNN models using a stochastic gradient method with a learning rate of 0.02. The error rate for each training epoch decreased as training proceeded (Fig. 5). When the error rate curve was saturated, we changed the learning rate from 0.02 to 0.001 for fine tuning. To avoid the overfitting problem, we chose the epoch with the lowest error rate in the validation set, even before the error rate curve of the training set became saturated (called, "early stopping"). We evaluated the accuracy of the optimized model for each fold in classifying the test data set.

For the statistical evaluation of the classification performance of the three classifiers (gCNN, pCNN and SVM), we conducted a one-way analysis of variance (ANOVA) of the classification accuracies at 10 folds with Bonferroni correction as a post hoc.

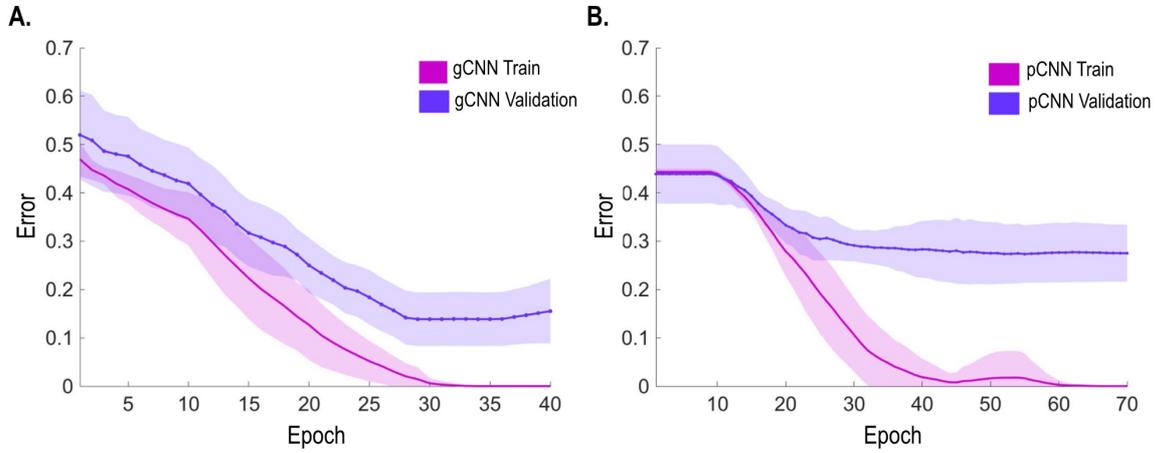

**Fig. 5.** Error rate learning curves along the training and validating epochs. The solid line represents average error rate of 10 models (for 10-folds), and the shaded region represents standard deviation of the models at each epoch. (A) and (B) show learning curves of gCNN and pCNN. gCNN was trained earlier and had superior performance over pCNN.

*Position invariance of local features*

As a successor of CNN, gCNN may have the position invariance property of detecting local features in the brain—a lower-level feature detector can often be used regardless of global position. To show this position invariance property, we conducted the same classification steps described above after globally rotating the spherical maps 45° and 90° (Fig. 6). Instead of retraining all layers in gCNN, we reused weights in the trained model up to 20 layers (five sets, each comprising a complex of convolution layer, batch layer, ReLU layer, and pooling layer, Nos. 1–20 of Table 1 in the Appendix) out of 25 layers from the bottom. Only the four upper layers of the model (a batch normalization layer, a ReLU layer, a hidden layer, and an output layer) were retrained with two rotated datasets. In this evaluation, we did not conduct fine-tuning of the reused layers. Similarly, we reused model weights up to 15 layers out of 19 layers in pCNN (Nos. 1–15, Table 2 in the Appendix). The levels of gCNN and pCNN reused in this evaluation were chosen before the fully connected softmax-classifier set. By reusing lower layers, the training time was significantly reduced.

In the two-dimensional projection, there were severe distortions in the local features, particularly in the area of the poles (Fig. 6B). Two-dimensional projections also led to discontinuity in the boundaries (0° or 360° in longitude, -90° or 90° in latitude). Although we tried to rectify this weakness by padding regions across the circular boundary, this may not have been sufficient. We surmised that the local features might not be maintained after global rotation in the 2D projected image. We compared the performances of gCNN and pCNN after rotation. The details of the model structures for gCNN and pCNN are presented in Tables 1 and 2 in the Appendix.

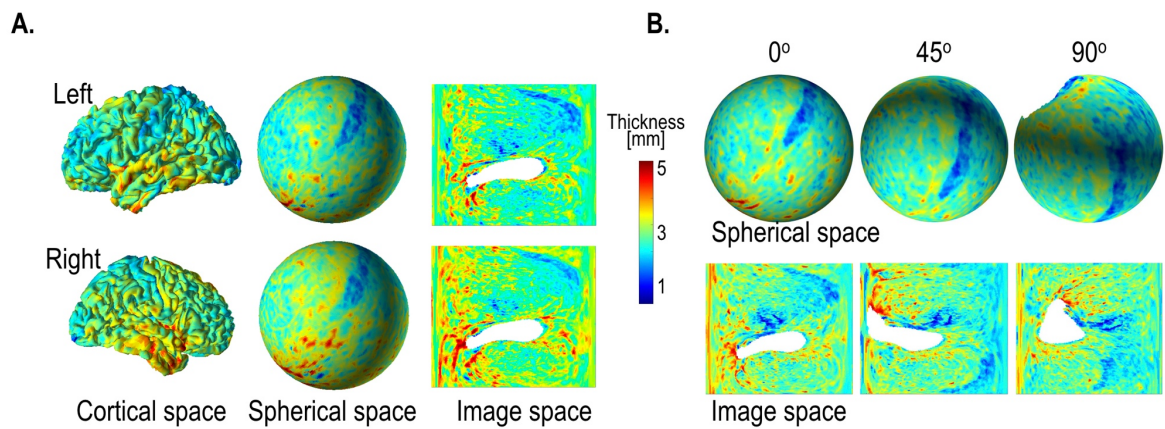

**Fig. 6.** An example of cortical thickness representation over various geometry and thickness representation after global rotation (45° and 90°). (A) Cortical thickness defined in the cortical surface can be transformed into spherical space and two-dimensional image space. The thickness map in the two-dimensional image was generated by projecting the spherical thickness map onto the image space based on latitude and longitude. (B) Spherical rotation of cortical thickness map changes representation on the sphere and images; the original, 45° rotation, and 90° rotated spheres and their projected images are presented. The global rotation

considerably influences the projected images, while the local pattern in the spherical map is maintained after rotation. The white areas indicate where thickness measurement is not available.

**Results**

Fig. 7. displays the group-average and group-differential surface-based cortical thickness maps used in the current study.

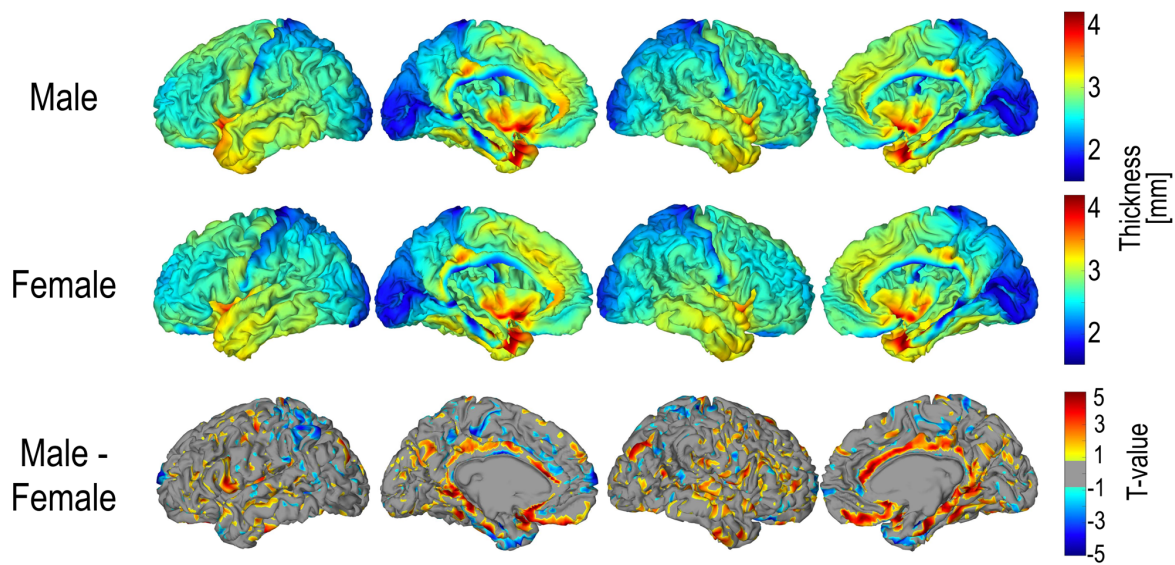

**Fig. 7.** Mean cortical thickness maps for male (N = 328) and female (N = 405) and their statistical difference. For statistical difference, only areas with p < 0.05 are presented. Areas that were thicker in male are represented in red and female in blue.

The classification performance of gCNN, SVM, and pCNN are summarized in Fig. 8. The one-way ANOVA of the classification accuracy showed a significant main effect (group difference) F(2,27) = 4.472 (p = 0.021). The average classification accuracy of gCNN was 87.14% (standard deviation (std) = 4.42), which is significantly higher than that of SVM (mean = 82.84%, std = 2.91, corrected p = 0.044) and pCNN (82.54%, std = 3.58, corrected p = 0.047). PCNN and SVM showed no significant difference in the classification accuracy.

*Classification performance of the rotated model*

In order to evaluate the position invariance of the local feature detection, we chose a $7^{th}$ fold model (red point in Fig. 8B) of gCNN as an initial model because the accuracy of that fold gCNN model (82.86%) was close to the mean accuracy of pCNN (82.54%). Using outputs at the 20-th layer in gCNN models (or the 15-th layer in pCNN models) for the input thickness data, we conducted 10-fold cross-validation of the remaining upper layers with a softmax classifier. The average accuracies for gCNN after rotation were 81.3% (std = 4.98) and 81.8% (std = 3.06) for 45° and 90° rotations, respectively, as shown in Fig. 8B. This accuracy after rotation is not a significant decrease from the initial model accuracy of 82.86%. On the other hand, pCNN showed average accuracies of 69.6% (std = 12.20) and 48.7% (std = 3.59) for the 45° and 90° rotations, which are significantly lower than the original accuracy (82.89%, std = 3.58), or within a chance level after 90° rotation.

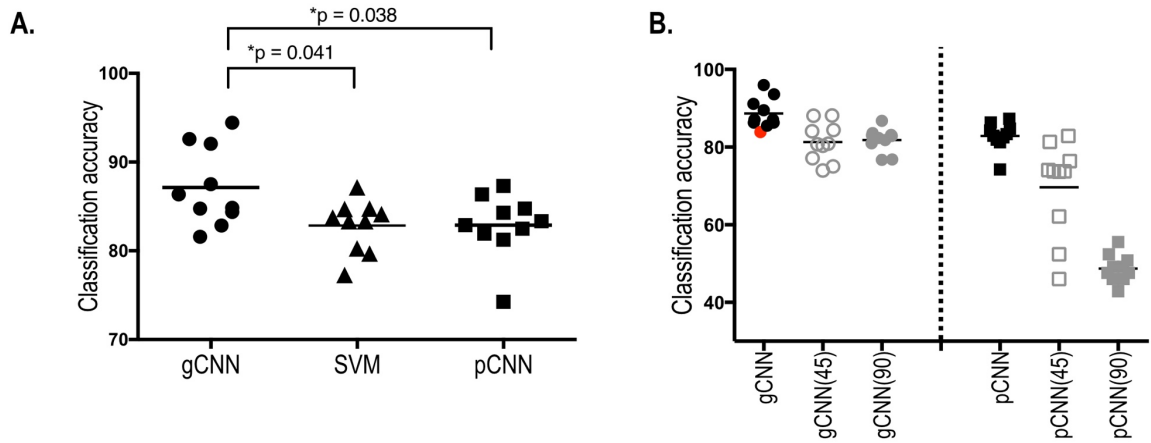

**Fig. 8.** Sex classification accuracies of gCNN, SVM, and pCNN and classification accuracy changes after global rotation. (A) When classification accuracy was evaluated using 10-fold cross-validation, a significantly superior performance was found in gCNN (mean = 87.14%, std = 4.42) compared to SVM (82.84%, std = 2.91, p = 0.038) and pCNN (82.54%, std = 3.58, p = 0.041) according to one-way ANOVA with Bonferroni correction as a post-hoc. (B) Classification accuracy did not significantly decrease after rotation in gCNN (45°: 81.27%, std = 4.98, 90°: 81.81%, std = 3.06) but significantly decreased in pCNN (45°: 69.64%, std = 12.2, 90°: 48.73%, std = 3.59) after rotation. The model with the red point fold (the seventh fold) was used as an initial (reused) model (82.85%) of gCNN for the position invariance test as it exhibited similar accuracy to pCNNs' mean accuracy.

## Discussion

In this paper, we proposed a geometric CNN method (gCNN) to deal with neuroimaging data on the cortical surface. To show its usefulness in surface-based representation, we applied gCNN to sex classification based on cortical thickness. The proposed method exhibited superior performance over the existing classification method (SVM) and conventional two-dimensional CNN (pCNN) for cortical thickness mapping. It also exhibited minimal performance deterioration after global shifts, which implies that the local features in gCNN are reusable.

*Surface-based methodology*

The human cortex has a surface geometry, which renders the representation of some neurometric features (e.g., cortical thickness) that are important for diverse neuroimaging researches. Surface representation is also advantageous in registering different brains for spatial normalization (Fischl et al., 1999b), registration between T1-weighted image and fMRI or diffusion weighted images (Greve and Fischl, 2009), efficient spatial smoothing, and partial volume correction of functional or metabolic imaging data (Greve et al., 2016; Greve et al., 2014; Park et al., 2006). Since these surface-based processing steps are efficient in preprocessing, removing statistical confounding factors and thus enhancing statistical power, surface-based analysis has widely been applied in diverse brain studies exploring morphometry (Landin-Romero et al., 2017), thickness (Goldman et al., 2009; Park et al., 2009; Rimol et al., 2012; Van Essen et al., 2017), myelination (Glasser and Van Essen, 2011; Van Essen et al., 2017), metabolic activity (Park et al., 2006), and tau and amyloid PET (Cho et al., 2016). These advantages of surface-based representation necessitate the development of a surface-based method for machine learning of cortical neuroimaging data.

*CNN and gCNN*

gCNN inherits the benefits of the CNN. Local filters in the convolution layer of CNN are characterized by sparse connectivity and shared weights across patches. A "replicated" local filter unit is effective in detecting common local features regardless of their position in image space. Furthermore, sharing weights increases learning efficiency by reducing the number of parameters that need to be trained. CNN also takes advantage of the hierarchical architecture, which entails multiscale information processing from local regions to global regions. This hierarchical and multiscale architecture for information abstraction is implemented by the pooling layer in the CNN. As a type of CNN, gCNN is efficient in detecting features hierarchically, which may explain the superior performance of gCNN over SVM and pCNN.

*Sex differences of cortical thickness*

To demonstrate the performance of gCNN in this study, we presented an example of a sex classification problem based on cortical thickness. In previous cortical thickness analyses of sex, several brain regions showed cortical thickness differences between males and females. For example, Sowell et al. (2007) reported significant cortical thinning in males compared to females at the right inferior parietal lobe and right posterior temporal regions and a tendency of thinning at the left ventral frontal and left posterior temporal regions. Other studies have shown increased cortical thickness in females compared to males in the frontal lobe and the parietal lobe (Allen et al., 2003; Nopoulos et al., 2000). Studies measuring gray matter density and cortical thickness have also shown local increases in gray matter in women, primarily in the parietal lobes (Good et al., 2001; Narr et al., 2005) and both the parietal and temporal lobes (Im et al., 2006; Luders et al., 2006). All of these studies were based on group data and showed diverse brain regions having different cortical thickness according to sex. Thus, it is not clear whether this finding can be applied to sex identification of each individual. In the current study, using 733 sets of data, gCNN utilized patterns of cortical thickness distribution to classify sex with reasonable accuracy.

The sex classification using cortical thickness may not be of a practical use that shows the clinical utility of this new machine learning algorithm. Nevertheless, the classification of sex based on the cortical thickness may be a good test-bed for different classifiers with a balanced data set (male and female). Indeed, the sex classification is not a trivial problem as reflected in the relatively low accuracy of conventional classification methods (e.g., less than 85% in SVM). Furthermore, the sex classification with a large-sized (HCP) database and a balanced number of class samples (e.g., man and woman) provides us a chance to train a gCNN model for the purpose of potential reuse of the model beyond sex-classification. In most classification studies with cortical thickness data, we may not have a sufficiently large number of data to train deep layers in gCNN. A gCNN model for sex-classification, if trained well using a large data set, may be reused in diverse applications, which is discussed in following section.

*Position invariance in gCNN*

In the hierarchical architecture of gCNN (as well as CNN), the lower-level filters are considered to detect features that are common to diverse applications, whereas global features in higher levels are more specific to each application. Therefore, lower-level feature detectors, which require sufficient data to train, can be reused for applications other than the purpose for which the model was trained, without significant distortions in the final outcome. After reusing the lower-level feature detectors, only the upper layer may be trained for the data with new application. This could not only reduce the computational cost but also mitigate the problem of insufficient data. We partially demonstrated this problem by shifting the global positions of cortical thickness while maintaining local properties. Compared to conventional CNN with two-dimensionally projected thickness image, gCNN shows a consistent level of accuracy after global rotations of thickness maps. This position invariance test suggests that we can reuse the lower-level feature detection of gCNN (found in the sex classification, for example) for the new applications, thus mitigating the need for a large number of samples for training from the beginning.

Instead of gCNN, one may consider two-dimensional projection of the cortical thickness for conventional CNN applications, as was found in EEG analysis (Bashivan et al., 2015). As shown in the rotation example, two-dimensional projection of 3D surface representation leads to inevitable distortions in representing common local features according to location. In particular, shapes near the pole are largely different from shapes in the equator. This violates the position invariance of the CNN in describing a common set of local features. The distortion during two-dimensional projection might be reduced by utilizing surface-based flattening (Fischl et al., 1999a); however, it still has problems associated with cutting surface into the two-dimensional image space. Flattening inevitably causes splits in the continuous representation, relocating close areas to distant areas, which may hinder model-reuse for different applications. Instead of requiring an additional flattening step, gCNN can be directly applied to the surface-based data using conventional CNN toolkits, with a slight modification.

*gCNN for surface-based representation*

The classification performance depends on how well the model is optimized. There are many factors that can be optimized in both gCNN and pCNN, such as model structure, training strategy and data augmentation, which should be chosen empirically depending on the application. The current study presented a gCNN example that exhibited superior performance compared to SVM and pCNN in sex classification, possibly owing to advantages inherent in the CNN, taking advantage of hierarchical feature detection and utilizing geometric information in the model without significant distortions.

Nevertheless, the purpose of the current study is not to show the general superiority of gCNN over the other methods as the performance may vary according to the quality of optimization. Instead, the current study is aimed at introducing a novel and simple CNN scheme that can easily be applied to surface-based or mesh-based neuroimaging data with a reliable accuracy.

gCNN differs from previous variants of 3D CNN in dealing with surface-based point data. In the classification of geometric shapes, previous studies transformed 3D points into voxels in the volume space, to which 3D CNN was applied. Examples include 3D ShapeNets (Wu et al., 2015), VoxeNet (using occupancy grid) (Maturana and Scherer, 2015), PointNet (Garcia-Garcia et al., 2016), and LightNet (Zhi et al., 2017). In contrast to voxel-type 3D applications, gCNN processes values (thickness in the current study) of the 3D points in the surface geometry while maintaining its geometry.

gCNN shares the basic mathematical framework of geodesic CNN formulated for geodesic shape classification on non-Euclidean manifolds (Boscaini et al., 2015; Bronstein et al., 2016; Masci et al., 2015). Although gCNN captures the pattern of feature (thickness) distribution in the same non-Euclidean space, if not the pattern of geometric shape features, gCNN can be considered as a specific type of non-Euclidean CNN. The advantage of gCNN over previous non-Euclidian CNNs (requiring complex geometric processing steps) is its technical simplicity achieved by a data sampling and reshaping method over a sphere, which is easily applicable to conventional software toolboxes implemented on GPUs without signigficant modification of the source code.

In addition to cortical thickness data, gCNN can be extended to various applications with diverse neuroimaging data such as fMRI or PET image after surface mapping. gCNN can also be used with a multi-layered input, concatenated with different modality or different cortical (three or six) layer information over the same cortical surface architecture.

In conclusion, gCNN takes advantage of both the CNN properties stemming from its brain-like architecture and surface-based representation of cortical information. gCNN may expedite surface-based machine learning in both scientific and clinical applications.

**Acknowledgement**

This work was supported by a grant from the National Research Foundation of Korea (NRF), funded by the Korean government (MSIP) (No. 2014R1A2A1A10052762).

**Appendix A.**

Table 1. gCNN architecture and parameter size

| No. | Type | Input size | Patch size/stride | Output size | Param. mem. |
|---|---|---|---|---|---|
| 1 | Convolution 1 | 40962x25x2 | 1x25 / 1 | 40962x1x36 | 7 KB |
| 2 | Batch normalization 1 | | | 40962x1x36 | 576 B |
| 3 | ReLU 1 | | | 40962x1x36 | |
| 4 | Mean pool 1 | 40962x1x36 | 1x6 or* 1x7 / 1 | 10242x1x36 | |
| 5 | Convolution 2 | 10242x25x36 | 1x25 / 1 | 10242x1x36 | 127 KB |
| 6 | Batch normalization 2 | | | 10242x1x36 | 576 B |
| 7 | ReLU 2 | | | 10242x1x36 | |
| 8 | Mean pool 2 | 10242x1x36 | 1x6 or 1x7 / 1 | 2562x1x36 | |
| 9 | Convolution 3 | 2562x25x36 | 1x25 / 1 | 2562x1x36 | 127 KB |
| 10 | Batch normalization 3 | | | 2562x1x36 | 576 B |
| 11 | ReLU 3 | | | 2562x1x36 | |
| 12 | Mean pool 3 | 2562x1x36 | 1x6 or 1x7 / 1 | 642x1x36 | |
| 13 | Convolution 4 | 642x25x36 | 1x25 / 1 | 642x1x36 | 127 KB |
| 14 | Batch normalization 4 | | | 642x1x36 | 576 B |
| 15 | ReLU 4 | | | 642x1x36 | |
| 16 | Mean pool 4 | 642x1x36 | 1x6 or 1x7 / 1 | 162x1x36 | |
| 17 | Convolution 5 | 162x25x36 | 1x25 / 1 | 162x1x36 | 127 KB |
| 18 | Batch normalization 5 | | | 162x1x36 | 576 B |
| 19 | ReLU 5 | | | 162x1x36 | |
| 20 | Mean pool 5 | 162x1x36 | 1x6 or 1x7 / 1 | 42x1x36 | |
| 21 | Fully connected layer 6 | 42x1x36 | 42x1 / 1 | 1x1x50 | 295 KB |
| 22 | Batch normalization 6 | | | 1x1x50 | 800 B |
| 23 | ReLU 6 | | | 1x1x50 | |
| 24 | Fully connected layer 7 | 1x1x50 | 1x1 / 1 | 1x1x2 | 408 B |
| 25 | Softmax classifier | 1 x 1 x 2 | | | |
| | Total | | | | 1.63MB |

* The size of mean pool at each node differs according to numbers of neighborhood nodes at the node, either 6 or 7.

Table 2. pCNN architecture and parameter size

| No | Type | Input size | Patch size/stride | Padding | Output size | Param. mem. |
|---|---|---|---|---|---|---|
| 1 | Convolution 1 | 224x224x2 | 11 x 11/ 4 | 1 | 54 x 54 x 64 | 61 KB |
| 2 | ReLU 1 | 54 x 54 x 64 | | | 54 x 54 x 64 | |
| 3 | Normalization 1 | 54 x 54 x 64 | | | 54 x 54 x 64 | |
| 4 | Mean pool 1 | 54 x 54 x 64 | 2 x 2 / 2 | 0 | 27 x 27 x 64 | |
| 5 | Convolution 2 | 27 x 27 x 64 | 5 x 5 / 1 | 2 | 27 x 27 x 64 | 400 KB |
| 6 | ReLU 2 | 27 x 27 x 64 | | | 27 x 27 x 64 | |
| 7 | Normalization 2 | 27 x 27 x 64 | | | 54 x 54 x 64 | |
| 8 | Mean pool 2 | 27 x 27 x 64 | 2 x 2 / 2 | 0 | 13 x 13 x 64 | |
| 9 | Convolution 3 | 13 x 13 x 64 | 3 x 3 / 1 | 1 | 13 x 13 x 64 | 144 KB |
| 10 | ReLU 3 | 13 x 13 x 64 | | | 13 x 13 x 64 | |
| 11 | Convolution 4 | 13 x 13 x 64 | 3 x 3 / 1 | 1 | 13 x 13 x 64 | 144 KB |
| 12 | ReLU 4 | 13 x 13 x 64 | | | 13 x 13 x 64 | |
| 13 | Convolution 5 | 13 x 13 x 64 | 3 x 3 / 1 | 1 | 13 x 13 x 64 | 144 KB |
| 14 | ReLU 5 | 13 x 13 x 64 | | | 162x1x36 | |
| 15 | Mean pool | 13 x 13 x 64 | 2 x 2 / 2 | | 6 x 6 x 64 | |
| 16 | Fully connected layer 6 | 6 x 6 x 64 | 6 x 6 / 1 | 0 | 1 x 1 x 100 | 900 KB |
| 17 | ReLU 6 | 1 x 1 x 100 | | | 1 x 1 x 100 | |
| 18 | Fully connected layer 7 | 1 x 1 x 100 | 1 x 1 / 1 | | 1 x 1 x 2 | 408 B |
| 19 | Softmax classifier | 1 x 1 x 2 | | | | |
| | Total | | | | | 1.79 MB |